\documentclass[times,twocolumn,final,authoryear]{elsarticle}

\usepackage{prletters}
\usepackage{framed,multirow}

\usepackage{amssymb}
\usepackage{latexsym}

\usepackage{url}
\usepackage{xcolor}
\definecolor{newcolor}{rgb}{.8,.349,.1}

\usepackage{caption}
\usepackage{subcaption}

\usepackage{dirtytalk}

\usepackage{adjustbox}

\usepackage{graphicx}

\setcitestyle{square} 

\journal{Pattern Recognition Letters}

\begin{document}

\ifpreprint
  \setcounter{page}{1}
\else
  \setcounter{page}{1}
\fi

\begin{frontmatter}

\title{Deep Network Pruning: A Comparative Study on CNNs in Face Recognition}

\author[1,2]{Fernando \snm{Alonso-Fernandez}\corref{cor1}}
\author[1]{Kevin \snm{Hernandez-Diaz}}
\author[2]{Jose Maria \snm{Buades Rubio}}
\author[1]{Prayag \snm{Tiwari}}
\author[1]{Josef \snm{Bigun}}

\cortext[cor1]{Corresponding author:
  Fernando Alonso-Fernandez (feralo@hh.se). Accepted at \textit{Pattern Recognition Letters}.}

\address[1]{School of Information Science, Computer and Electrical Engineering, Halmstad University, Box 823, Halmstad SE 301-18,
Sweden}

\address[2]{Computer Graphics and Vision and AI Group, University of Balearic Islands, Spain}


\begin{abstract}
The widespread use of mobile devices for all kinds of transactions makes necessary reliable and real-time identity authentication, leading to the adoption of face recognition (FR) via the cameras embedded in such devices.
%
%
Progress of deep Convolutional Neural Networks (CNNs) has provided substantial advances in FR. Nonetheless, the size of state-of-the-art architectures is unsuitable for mobile deployment, since they often encompass hundreds of megabytes and millions of parameters.
We address this by studying methods for deep network compression applied to FR.
In particular, we apply network pruning based on Taylor scores, where less important filters are removed iteratively.
The method is tested on three networks based on the small SqueezeNet (1.24M parameters) and the popular MobileNetv2 (3.5M) and ResNet50 (23.5M) architectures.
These have been selected to showcase the method on CNNs with different complexities and sizes.
We observe that a substantial percentage of filters can be removed with minimal performance loss. 
Also, filters with the highest amount of output channels tend to be removed first, suggesting that high-dimensional spaces within popular CNNs are over-dimensioned.

\end{abstract}

\begin{keyword}
\KWD Face Recognition\sep Mobile Biometrics\sep Network Pruning\sep Taylor Expansion\sep Deep Learning\sep Convolutional Neural Networks

\end{keyword}

\end{frontmatter}


\section{Introduction}

Mobile biometrics can offer secure, user-friendly authentication for many services such as e-commerce, banking, messaging, remote work, social media, education, healthcare, government services, etc.
Here, we address face recognition (FR), where Convolutional Neural Networks (CNN), as in many other vision tasks, has become the popular tool \citep{[Sundararajan18-DLbiometrics]}.
Given enough data, CNNs produce classifiers with impressive performance in
unconstrained scenarios with high variability.
However, the limited computational resources of mobile devices make that popular CNNs
are often unsuitable due to their complexity and large model sizes. 
This creates the challenge of developing
lighter and more efficient models that maintain accuracy without heavy resource requirements.

There is, therefore, interest in developing light biometric CNNs, driven by the need for 
authentication on devices with limited processing.
Here, we apply network compression \citep{caldeira2024arxiv_model_compression_biometrics} to existing popular CNN architectures to create more compact models without sacrificing accuracy.
We use a pruning method based on importance scores of network filters \citep{Molchanov19CVPRcnnPruningTaylor}, which quantifies the impact on the network error if filters are removed. Such scores are obtained via first-order Taylor approximation, which only requires the gradient elements from backpropagation.

In an earlier work \citep{Alonso23IWAIPR_SqueezerFaceNet}, we used this method to reduce an already small SqueezeNet network of 1.24M parameters \citep{[Iandola16SqueezeNet],[Alonso20SqueezeFacePoseNet]}.
Such work was among the first ones to evaluate network pruning for FR \citep{caldeira2024arxiv_model_compression_biometrics}.
This paper extends the study to MobileNetv2 \citep{[Sandler18mobilenetv2]} (3.5M parameters) and ResNet50 \citep{[He16]} (23.5M).
Such architectures are selected to consider different model sizes.
The purpose and contributions of this work are, therefore, multi-fold.
We first summarize the literature on network compression, in particular pruning for FR.
Then, we study the impact of the employed pruning method on the FR performance of three different-sized CNN architectures. 
In our experiments, we prune up to 99\% of the filters, maintaining performance until a certain percentage.
SqueezeNet and MobileNetv2 allow up to 20-30\% pruning, situating them at $\sim$1M parameters.
The larger ResNet50 allows up to 99\% pruning in some cases, which, interestingly, also corresponds to $\sim$1M parameters.
We also analyze the effect of pruning in network layers, observing that those with most channels are pruned first.

\section{Related Works}
\label{sec:soa}

Several light CNNs have been presented across the years \citep{[Iandola16SqueezeNet],[Sandler18mobilenetv2],[Zhang18ShuffleNet],Tan19BMVC_MixNets,zhang2020vargnet},
%
%
%
%
%
%
%
%
mainly for generic visual tasks in the context of ImageNet \citep{[Russakovsky15_ImagenetChallege]}, 
using techniques that allow faster processing and fewer parameters, such as point-wise convolution, depth-wise separable convolution, variable group convolution, mixed convolution, channel shuffle, and bottleneck layers.
Some have been used for FR,
either in their original form or reducing channels or layers for compactness 
\citep{[Chen18MobileFaceNets],[Duong19MobiFace],[Martinez19ShuffleFaceNet],Yan2019ICCVW_VarGFaceNet,[Alonso20SqueezeFacePoseNet],Boutros2021IJCB_MixFaceNets}.
In doing the latter, however, the number of channels or layers is reduced heuristically just to obtain a more compact network.
%
%
%
%
%
%
%
%
%
%
%
%
%
%
Some other works have specifically addressed optimizations for low-power devices, such as considering mobile face datasets \citep{Cimmino22access_mobile_masked_FR},
hierarchical clustering to skip convolutions in image blocks that are similar to each other \citep{Kim20TCS_CNN_accelerator_FR_mobile}, 
using several shallow CNNs to process each face region separately \citep{Abate23prl_ablation_part_FR},
or applying re-lightning and pose correction techniques to fine-tune the recognition network and achieve increased performance in difficult mobile data \citep{Yao22access_FR_low_power_edge_AI}.
%
%
%
%
%

Instead of manual adaptation, the work \citep{boutros2021Access_Pocketnet} suggested Neural Architecture Search (NAS) to design light FR models named PocketNets.
Another increasingly used strategy is network compression. 
Most biometrics-related compression works have focused on FR, with some others applied to ocular recognition \citep{Rattani23ACCESS_OcularCNNPruningBenchmark}, face detection \citep{Jiang22CCS_PruneFaceDet}, and, to a minor extent, iris, expression, emotion, morphing attacks, or sclera segmentation \citep{caldeira2024arxiv_model_compression_biometrics}.
Compression techniques include quantization, knowledge distillation (KD) and pruning.
Model quantization converts full-precision (FP) weights and activations (typically in 32 bits) to a lower precision (LP) of 8 bits to up to 1 bit, maintaining the same architecture.
LP is less computationally expensive, and several hardware and libraries are optimized for fast 8-bit processing.
%
%
KD uses a teacher network, usually a complex model that performs well in the proposed task, to guide a lighter student model.
The student is trained to produce the teacher's output via loss functions that minimize the gap between their responses.
%
%
%
%
However, as with quantization, the architecture of KD networks is fixed. 
In network pruning, on the other hand, redundant parts 
are eliminated, resulting in a more sparse network.
To do so, network weights are usually ordered according to some criteria, and those with the lowest importance are removed until the desired level of sparsity. 
%
%

Most network compression literature for FR and biometrics uses KD \citep{caldeira2024arxiv_model_compression_biometrics}.
Apart from our mentioned work \citep{Alonso23IWAIPR_SqueezerFaceNet}, 
few have applied pruning \citep{Polyak15access_facepruning,Liu22pami_facepruning}.
The paper
\citep{Polyak15access_facepruning} 
proposed two methods to remove low contributing channels:
inbound pruning (IP) and reduce and reuse pruning (RRP).
%
%
%
IP prunes inbound channels of a layer, whereas RRP prunes outbound channels.
Important channels are determined to be those with a high variance in their activations. Thus, connections with low variance are removed. 
%
%
%
%
%
%
%
%
%
%
The methods were tested on CASIA and LFW datasets, improving runtime up to 2.65 with less than a 2\% accuracy drop.
The work \citep{Liu22pami_facepruning} 
proposed to remove channels 
using a loss function with two terms: a reconstruction error between the feature maps of the baseline network and the pruned one,
and a discrimination-aware term that uses the output of intermediate layers as features to do the actual softmax classification.
%
%
FR experiments used LResNet34E-IR and Mobile-FaceNet as backbones (trained on MS-Celeb-1M) and four well-known benchmark datasets (LFW, CFP-FP, AgeDB-30, and MegaFace), showing negligible performance drops after pruning 25-50\% of the channels.

\section{Materials and Methods}

\subsection{Network Pruning}
\label{sect:pruning_method}

We apply the method for general network pruning of \citep{Molchanov19CVPRcnnPruningTaylor}.
It iteratively estimates the importance scores of individual elements based on their effect on the network loss.
Then, elements with the lowest scores are pruned. 
Given a network with parameters
$\textbf{W}=\left\{ w_{0},w_{1},...,w_{M} \right\}$ and a training set $\mathcal{D}$ of input $\left( x_{i} \right)$ and output $\left( y_{i} \right)$ pairs
$\mathcal{D}=\left\{ \left( x_{0},y_{0} \right),\left( x_{1},y_{1} \right),...,\left( x_{K},y_{K} \right) \right\}$,
the aim of network training is to minimize the classification error $E$ by solving


\begin{equation}
 \mathop {\min }\limits_{\textbf{W}} E(\mathcal{D},{\textbf{W}}) =
 \mathop {\min }\limits_{\textbf{W}} E(y|x,{\textbf{W}})
\end{equation}

The importance of a parameter $w_{m}$ can be defined by its impact on the error if it is removed.
Under an \textit{i.i.d.} assumption, the induced error can be quantified as: 

\begin{equation}
\label{eq:induced_error}
    {\mathcal{I}}_m=\bigg( E\left( \mathcal{D},\textbf{W} ) - E( \mathcal{D},\textbf{W}|w_{m}=0 \right)  \bigg)^{2}
\end{equation}

Computing $\mathcal{I}_m$ for each parameter 
would demand to evaluate $M$ versions of the network. 
%
This is avoided by approximating $\mathcal{I}_m$ in the vicinity of $\textbf{W}$ by its first-order Taylor expansion:


\begin{equation}
\label{eq:taylor_expansion}
    {\mathcal{I}}_{m}^{1}(\textbf{W})=\left( g_{m}w_{m} \right)^{2}
\end{equation}

\noindent where $g_{m}=\frac{\partial E}{\partial w_{m}}$ are the elements of the gradient $g$. A second-order expansion is possible via the Hessian of $E$ \citep{Molchanov19CVPRcnnPruningTaylor}, but we employ the first-order approximation for faster computation.
The gradient $g$ is available from backpropagation, so $\mathcal{I}_{m}^{1}$ can be easily computed.
The joint importance of a set of parameters $\textbf{W}_S$ (e.g. a filter) can then be obtained as:

\begin{equation}
\label{eq:joint_induced_error}
    {\mathcal{I}}_{S}^{1}(\textbf{W}) \triangleq \sum_{s\in S}^{}\left( g_{s}w_{s} \right)^{2}
\end{equation}

The algorithm starts with a trained network, which is pruned iteratively over the same training set.
Given a mini-batch, the gradients are computed, the network weights are updated by gradient descent, and the importance of each filter is obtained via Eq.~\ref{eq:joint_induced_error}.
At the end of each epoch, the scores of each filter are averaged over the mini-batches, and those with the smallest importance are removed.
%
%
The resulting network can then be fine-tuned again 
to regain potential accuracy losses. 

\begin{table}[t]
\small
    \caption{Networks evaluated in this paper.}
    \centering
    \begin{tabular}{cccccc}
    \textbf{} &  \textbf{Conv} &  \textbf{Total} & \textbf{Model} & \textbf{Para-} & \textbf{Vector} \\

    \textbf{Model}  & \textbf{Layers} & \textbf{Filters} & \textbf{Size} & \textbf{meters} & \textbf{Size} \\
    \hline \hline

SqueezeNet & 18 & 3168 & 4.6MB & 1.24M & 1000  \\

MobileNetv2 & 53 & 7950 & 13MB & 3.5M  & 1280  \\

ResNet50 & 50 & 21274 & 83.9MB & 23.5M  & 2048  \\ \hline

    \end{tabular}
    \label{tab:networks_used}
\end{table}

\begin{figure}[t]
\centering
        \includegraphics[width=.4\linewidth]{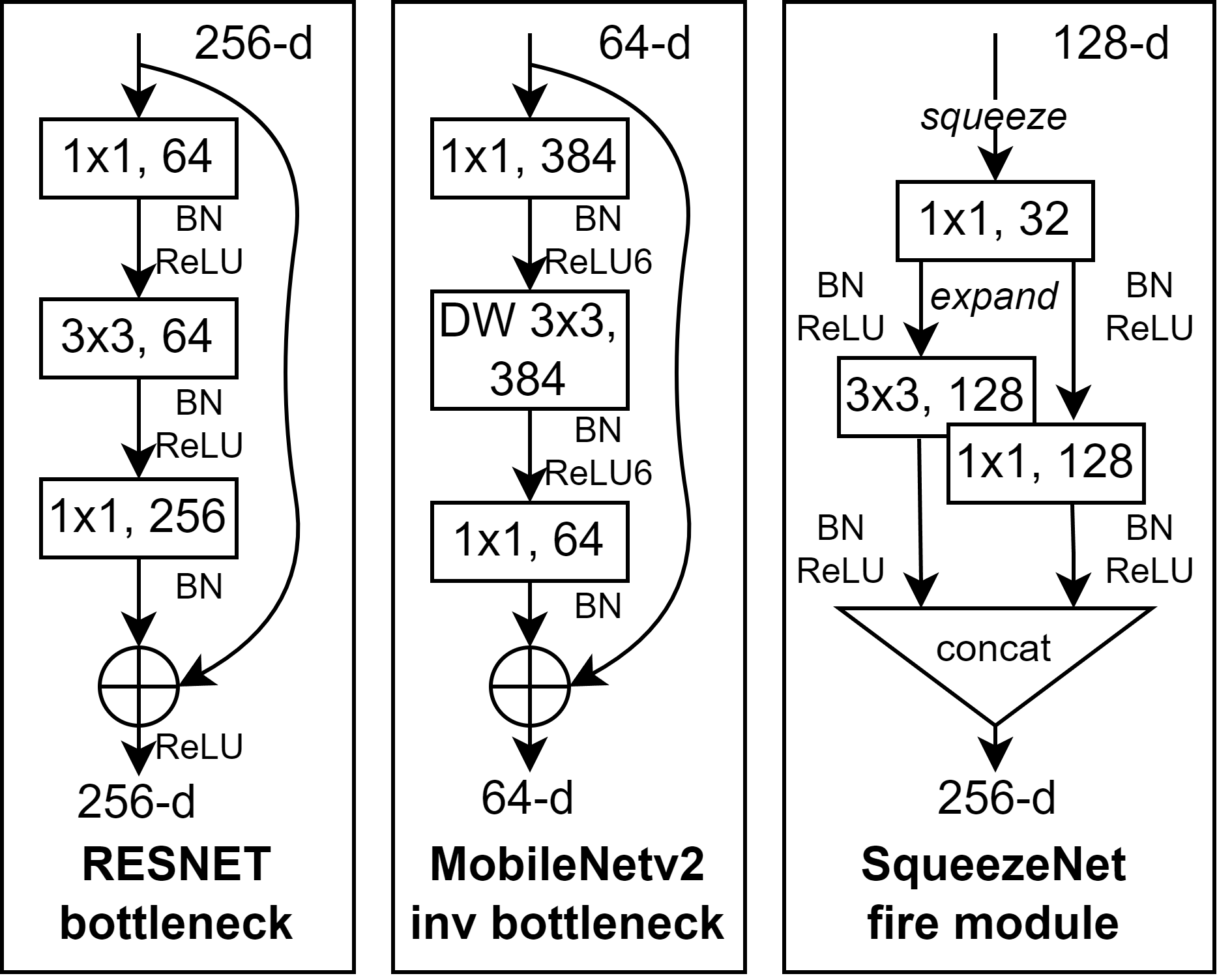}
\caption{Internal architecture of the building blocks of the CNN architectures employed. Inspired by \citep{[He16]}.
}
\label{fig:cnn-blocks}
\end{figure}

\begin{figure}[t]
\centering
        \includegraphics[width=.75\linewidth]{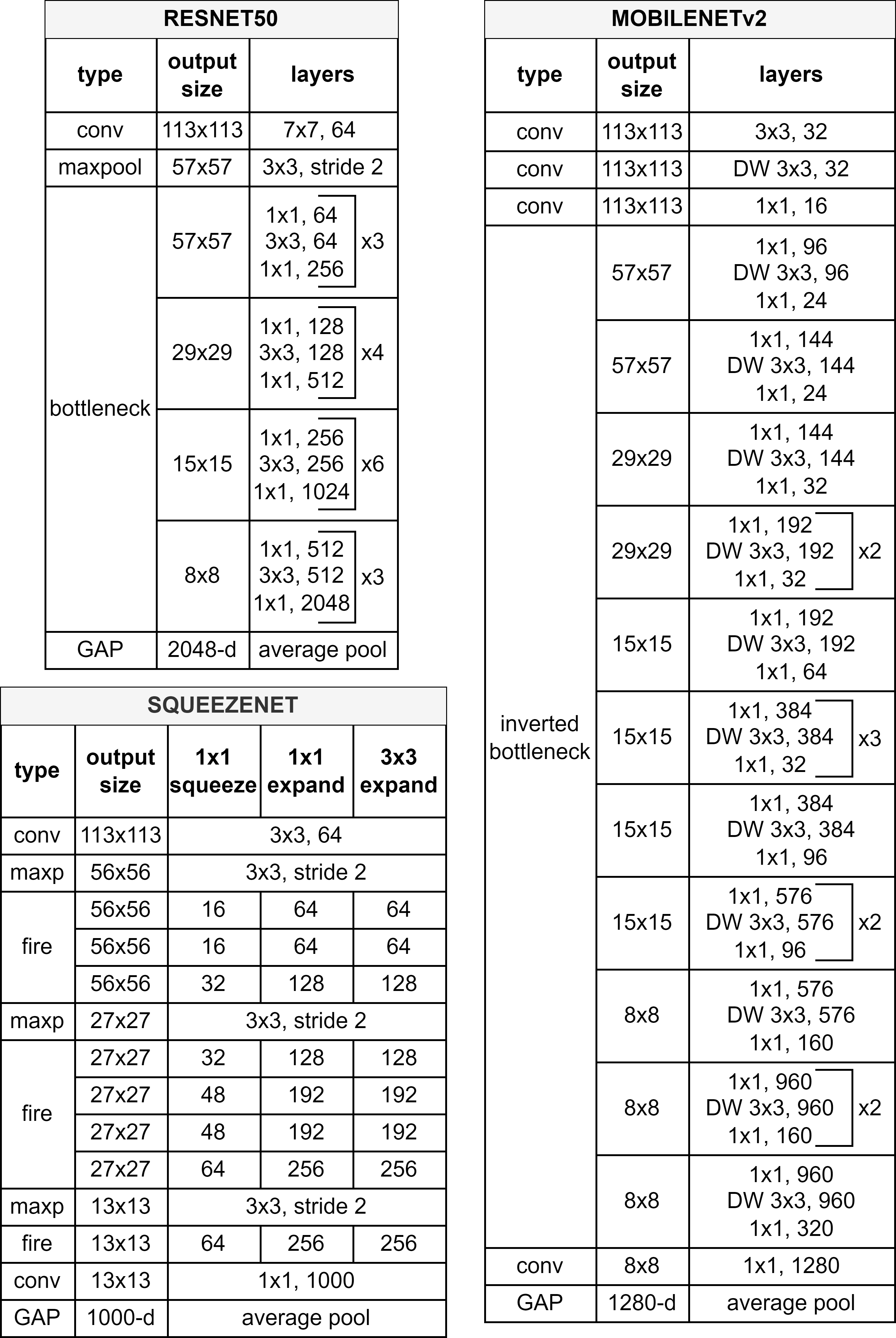}
\caption{Architecture of the networks used.}
\label{fig:cnn-structure}
\end{figure}

\subsection{Recognition Architectures}
\label{sect:cnns}

We use three CNNs, chosen to account for a different number of parameters or depth (Table \ref{tab:networks_used}).
They comprise two light architectures (SqueezeNet, MobileNetv2), and a large ResNet50.
%
\textit{ResNet} networks \citep{[He16]} introduced residual connections, allowing the input from a lower layer to reach a higher layer, bypassing intermediate ones.
Followed later by many (including MobileNet), this allows deeper networks and eases training by improving gradient propagation.
The authors also introduced bottleneck blocks (Figure~\ref{fig:cnn-blocks}, left) where high input dimensionality is first reduced with 1$\times$1 point-wise filters, then 3$\times$3 filters are applied over the lower dimensional space, and finally 1$\times$1 filters regain the input dimensionality. The larger 3$\times$3 filters are applied in the reduced dimensionality space, resulting in fewer parameters. We use the ResNet50 variant with 50 convolutional layers, 23.5M parameters, and 83.9 Mb.
%
%
%
\textit{MobileNetv2} \citep{[Sandler18mobilenetv2]} 
uses 
inverted residuals to achieve a light architecture (3.5M parameters, 13Mb and 53 convolutional layers).
In inverted residuals (Figure~\ref{fig:cnn-blocks}, centre), also called inverted bottleneck blocks with expansion, channel dimensionality is first expanded (instead of reduced) with 1$\times$1 filters.
Then, 3$\times$3 filters are applied, but to cope with the increased dimensionality, 
they are implemented via depth-wise separable convolution for parameter and processing savings.
Finally, channel dimensionality is reduced again with 1$\times$1 filters.
%
%
%
%
\textit{SqueezeNet} \citep{[Iandola16SqueezeNet]} is among the smallest architectures proposed within ImageNet (18 convolutional layers, 1.24M parameters, and 4.6 MB), and one of the early attempts to reduce the network parameters and size.
It uses \textit{squeeze} and \textit{expand} steps, similar to bottlenecks, but with a stack of two layers, called \textit{fire} module (Figure~\ref{fig:cnn-blocks}, right).
The channel dimensionality is first \textit{squeezed} with 1$\times$1 filters, and then \textit{expanded} with a larger amount of 3$\times$3 and 1$\times$1 filters. 
SqueezeNet uses late downsampling to keep larger activation maps as much as possible, and it is a sequential CNN without residuals.
%

Figure~\ref{fig:cnn-structure} shows the architecture of the networks from our training environment (Matlab r2023b), with slight changes.
They use a 113$\times$113 input by changing the stride of the first convolutional layer from 2 to 1.
This allows keeping the network unchanged and reusing ImageNet pre-trained networks as starting models for faster convergence \citep{[Kornblith19imagenet_transfer_better]}.
SqueezeNet uses ReLU, and we have added batch normalization (missing in the original implementation), which helps to improve performance \citep{Alonso23IWAIPR_SqueezerFaceNet}.
Bottleneck blocks in ResNet50 include batch normalization and ReLU. 
MobileNetv2 uses ReLU6 
(clipped ReLU), but only after the first two convolutions of inverted bottlenecks.


\subsection{Databases}
\label{sec:databases}

\begin{figure}[t]
\centering
        \includegraphics[width=.9\linewidth]{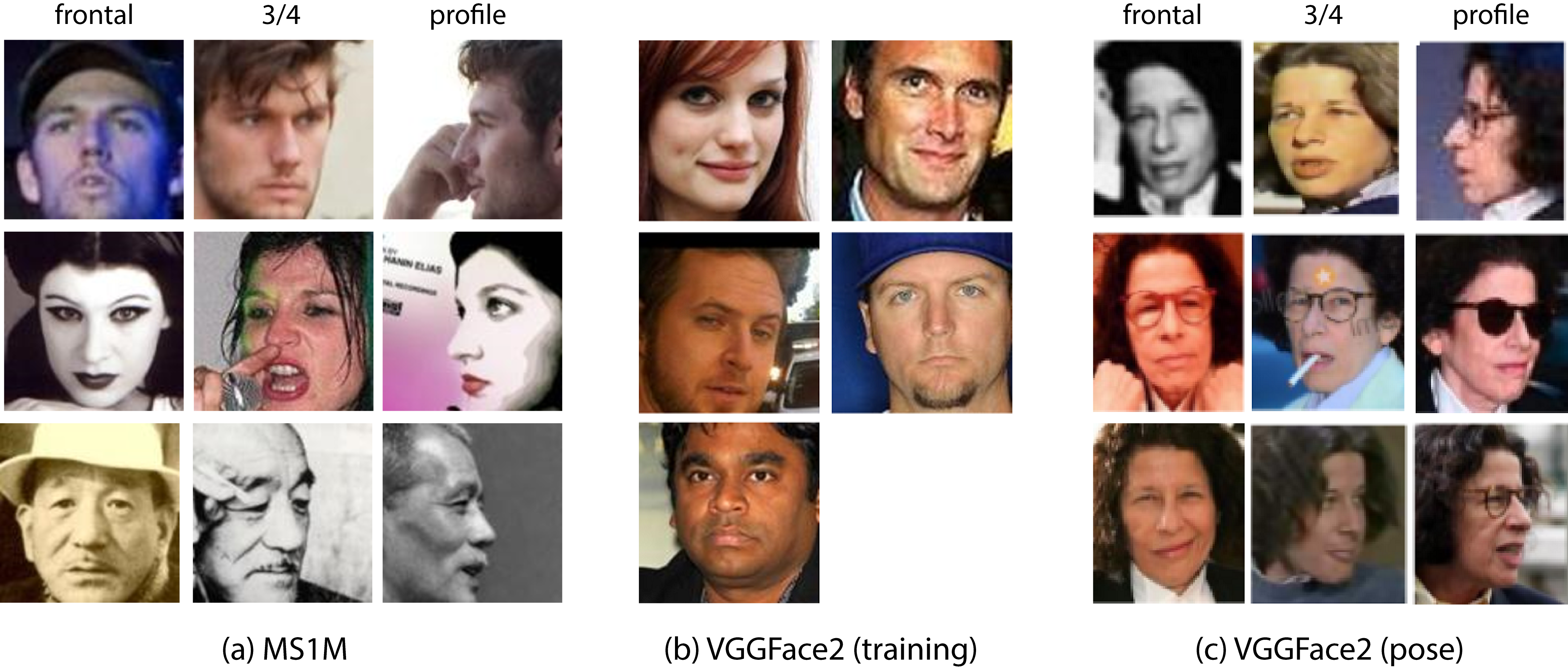}
\caption{Example images of the databases used. (a) MS1M from three users (by row) and three viewpoints (column). (b) VGGFace2 training images with a random crop. (c) VGGFace2 pose templates from 
three viewpoints (by column).
}
\label{fig:databases}
\end{figure}

VGGFace2 \citep{[Cao18vggface2]} is used for training and evaluation, having 3.31M images of 9131 celebrities (362.6 images per person on average). The images are sourced from the internet, with significant variations in pose, age, ethnicity, lighting, and background.
They are divided into 8631 training classes (3.14M images) and 500 testing classes. 
A subset of 368 subjects from the test set is defined, VGGFace2-Pose, with 10 images per pose (frontal, three-quarter, profile) and 11040 images.
To improve performance, we pre-train the networks with the RetinaFace cleaned subset of MS-Celeb-1M \citep{[Guo16_MSCeleb1M]} (referred to as MS1M), which includes 5.1M images of 93.4K identities. 
%
%
While MS1M has more images, its intra-identity variation is limited due to an average of 81 images per person.
We address this by pre-training on MS1M and then fine-tuning on the more diverse VGGFace2, which demonstrated superior performance compared to training on VGGFace2 only \citep{[Cao18vggface2],[Alonso20SqueezeFacePoseNet]}.
Some examples of the databases are shown in Figure \ref{fig:databases}.

\begin{table}[t]

\caption{Number of biometric verification scores. Same pose experiments include three types (which explains the 3$\times$ term): fontal vs. frontal, 3/4 vs. 3/4 and profile vs. profile. Similarly, cross-pose experiments include: frontal vs. 3/4, frontal vs. profile, and 3/4 vs. profile.}
\centering

\begin{adjustbox}{max width=.46\textwidth}


\begin{tabular}{ccccc}


\textbf{Template} & \textbf{Type} & \textbf{Same pose}  & \textbf{Cross pose} & \textbf{Total} \\ \hline

1-1  & Genuine  &  3 $\times$ 368 $\times$ (9+8+...+1)  & 3 $\times$ 368 $\times$ 10 $\times$ 10   &  160,080  \\

     & Impostor  &  3 $\times$ 368 $\times$ 100  & 3 $\times$ 368 $\times$ 100  &  220,800  \\ \hline

5-5  & Genuine  & 3 $\times$ 368 $\times$ 1  &  3 $\times$ 368 $\times$ 2 $\times$ 2  & 5,520     \\

     & Impostor  &  3 $\times$ 368 $\times$ 100   & 3 $\times$ 368 $\times$ 100    &  220,800 \\ \hline


\end{tabular}



\end{adjustbox}

\label{tab:scores}

\end{table}

%
%
%
%
%
%
%
%
%
%
%
%
%
%
%
%
%
%
%
%

\subsection{Training and Verification Protocols}
\label{sect:protocol}

The networks are trained for biometric identification, using ImageNet weights as initialization and soft-max as the activation function in the last layer.
Training and evaluation follow the VGGFace2 protocol  \citep{[Cao18vggface2]}.
During training, the shortest side of VGGFace2 bounding boxes is resized to 129 pixels, and then a random 113$\times$113 crop is taken.
%
%
MS1M images are directly at 113$\times$113. 
%
%
Additionally, we apply a horizontal random flip.
We use SGDM with a mini-batch of 128.
The learning rate starts at 0.01 and is reduced to 0.005, 0.001, and 0.0001 when the validation loss plateaus.
We reserve 2\% of images per user in the training set for validation.
MS1M users with fewer than 70 images are excluded to avoid
%
fully connected layers dedicated to under-represented classes, resulting in 35016 users and 3.16M images. 


We conduct verification experiments with VGGFace2-Pose resizing the shortest image side to 129 pixels, then taking a centre 113×113 crop.
%
%
%
Genuine (mated) scores are obtained by comparing each template of a user to his/her remaining ones, avoiding symmetric comparisons. For impostor (non-mated) scores, the first template of a user is 
compared with the second template of the next 100 users.
Identity templates are created by combining five faces of the same pose \citep{[Cao18vggface2]}, leading to two templates per user and pose.
To test the robustness of the network and the pruning under more challenging conditions, we also use single-image templates.
Template vectors are created by averaging face descriptors from the Global Average Pooling layer.
%
%
To combat pose variation, we average the descriptor of an image with its horizontally flipped counterpart \citep{[Duong19MobiFace]}. The cosine similarity is used for template comparison.
Table~\ref{tab:scores} shows the total number of scores. 

\begin{figure*}[htb]
    \centering
    \includegraphics[width=0.95\textwidth]{SQ_MB2_R50_performance_all.png}
    \caption{Face verification results on the VGG2-Pose database (EER \%) over the network pruning process. Data tips show selected values of the red curve (after pruning + retraining at 0.01).}
    \label{fig:results_EER_all}
\end{figure*}

\begin{figure}[htb]
    \centering
    \includegraphics[width=0.4\textwidth]{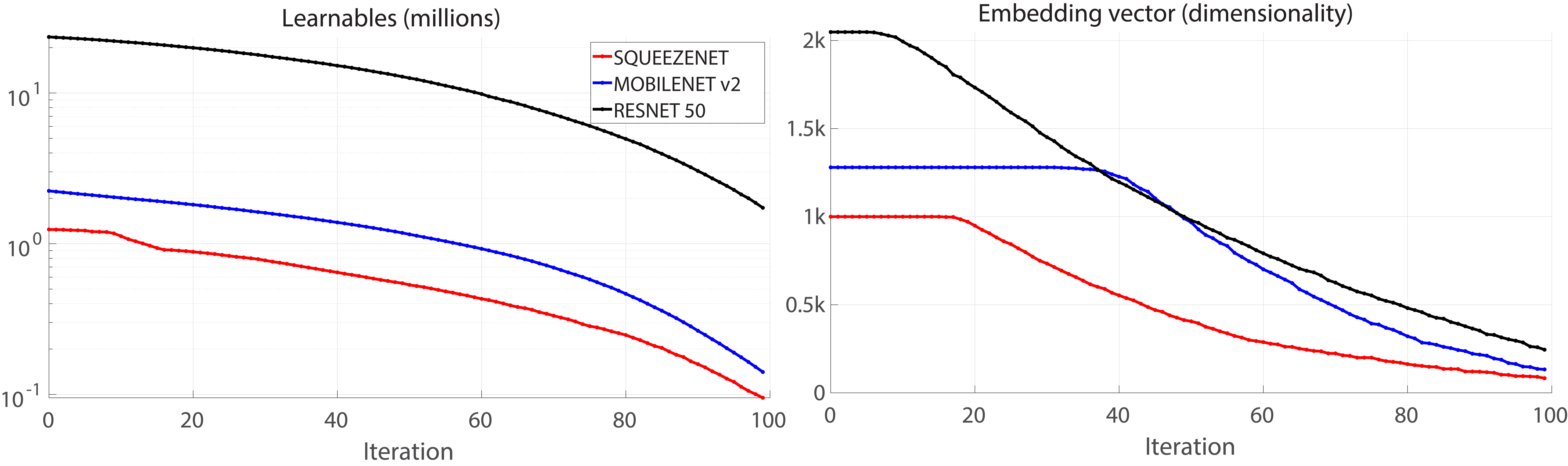}
    \caption{Effect of pruning in the network (learnables and embedding size).}
    \label{fig:results_network_param_size_embedding}
\end{figure}

\begin{figure}[htb]
    \centering
    \includegraphics[width=0.4\textwidth]{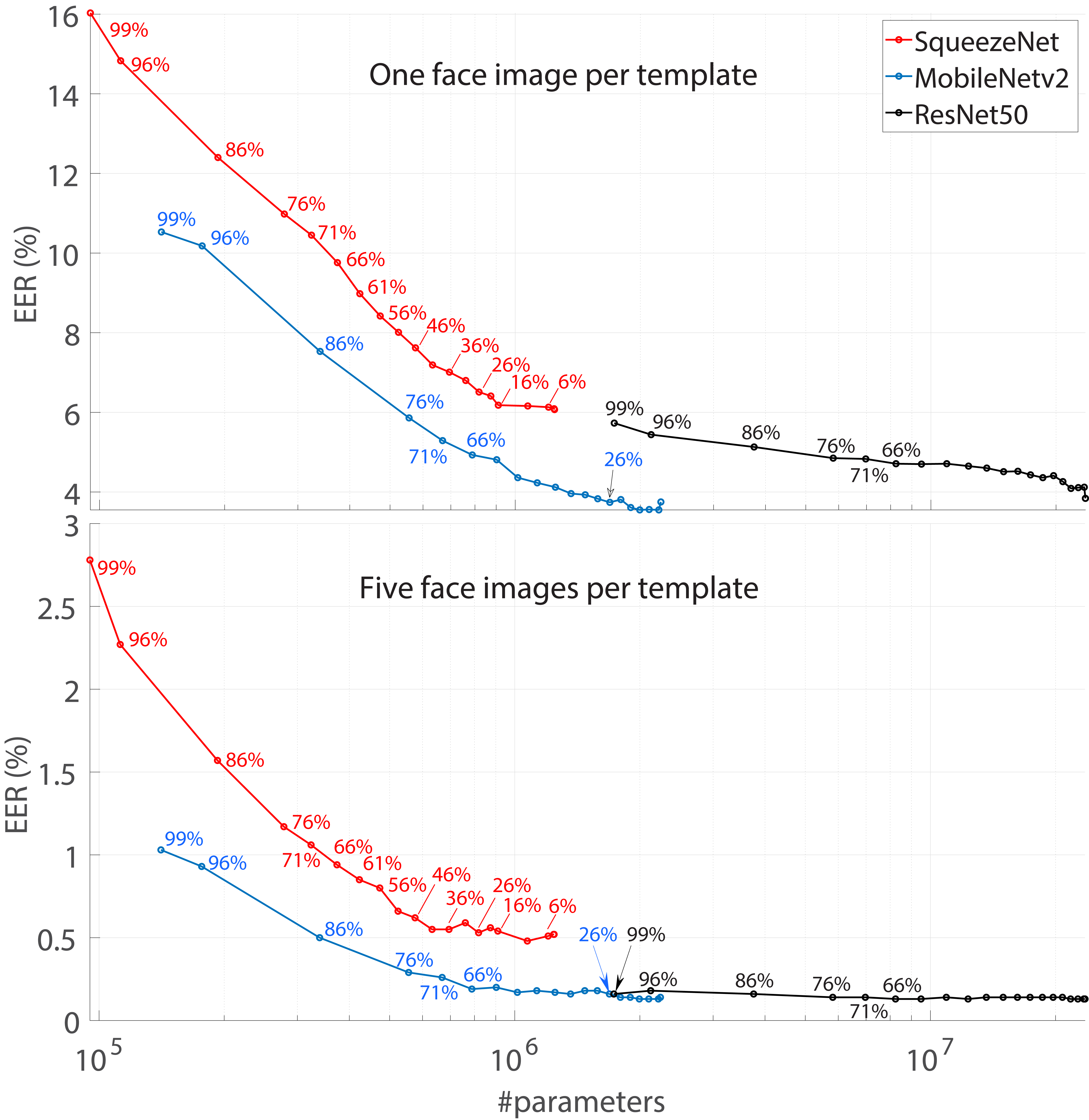}
    \caption{Verification accuracy on VGG2-Pose (ERR \%) vs. amount of learnables. The labels in the curves indicate the corresponding percentage of pruned filters.}
    \label{fig:results_performance_learnables}
\end{figure}

\begin{figure}[htb]
    \centering
    \includegraphics[width=0.45\textwidth]{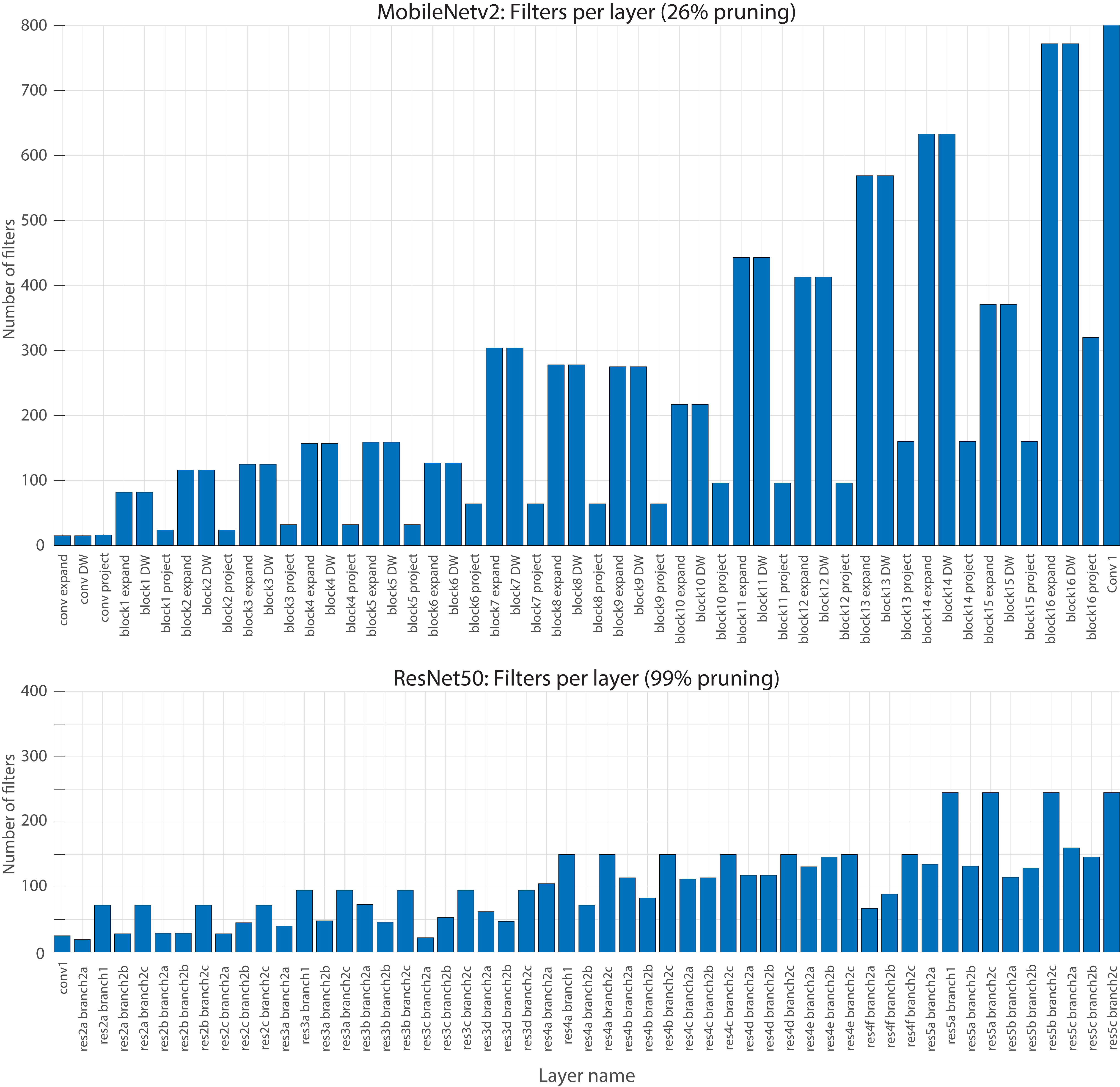}
    \caption{MobileNetv2 and ResNet50 structure after 26\% and 99\% pruning respectively. The layer names correspond to the networks' release of Matlab r2023b. The last layer of MobileNetv2 has 1280 filters (capped for visualization purposes). Better if zoomed.}
    \label{fig:_filters_per_layer_26pc_99pc_pruning}
\end{figure}

\section{Results and Discussion}
\label{sec:results}

We apply pruning to the networks using a random 25\% of the VGGFace2 training set to compute the filter importance scores.
We then remove 1\% of the lowest-scoring filters after each epoch.
The optimizer is SGDM with mini-batch 128 and a learning rate of 0.01.
Figure \ref{fig:results_EER_all} (blue curves) shows the EER of the pruned networks on VGGFace2-Pose. 
The baseline EER at $x$=0 (no pruning) is 6.07\% (SqueezeNet), 3.75\% (MobileNetv2) and 3.84\% (ResNet50) in one-to-one comparisons (single-image template), and 0.52\%, 0.14\% and 0.13\% for five-to-five comparisons.
The deeper MobileNetv2 and ResNet50 outperform the lighter SqueezeNet, with MobileNetv2 matching or exceeding ResNet50 with fewer parameters and size, demonstrating the effectiveness of the MobileNetv2 design to provide good accuracy with a reduced network.

After just removing 1\% of the filters (first iteration), a sharp increase in error rates occurs. We attribute this to overfitting since the networks have been trained previously over the same set with early stopping. 
%
After this jump, performance stabilizes or improves until a certain percentage of the filters are pruned, but the behaviour differs substantially per network.
SqueezeNet maintains performance until 15-20\% pruning (in five-to-five comparisons, until 40\%, albeit with some oscillations).
MobileNetv2, on the other hand, is very sensitive to filter removal, and performance drops substantially after 10\% pruning.
ResNet50 remains stable even after 60-70\% pruning, showing its robustness to filter removal.
However, even with 70\% filters removed, ResNet50 has more parameters than the other networks, as it will be analyzed later.

To regain accuracy losses, pruned networks are retrained again on VGGFace2. 
The results are given in Figure \ref{fig:results_EER_all} as well.
We use a starting learning rate of 0.01 (red curve) and 0.001 (orange).
We hypothesise that since the network has already been trained on the same database, starting with a high rate may be counterproductive.
However, this is not the case, since the best results are with a starting rate of 0.01.
%
%
All architectures maintain accuracy up to a certain pruning percentage.
In addition, using five images per template allows higher pruning sparsities. Combining several images may produce that the most relevant features of each identity are kept, while random information is averaged out \citep{caldeira2024arxiv_model_compression_biometrics}.
However, similarly to the above, the behaviour per network differs.
 SqueezeNet maintains performance until 15-20\% (one-to-one) or 20-30\% pruning (five-to-five). 
ResNet remains stable even with high pruning, and the performance gap w.r.t. the unpruned network is eliminated.
On the other hand, the sensitiveness of MobileNetv2 to pruning is substantially reduced after retraining, maintaining a stable performance until 20-30\% pruning in both one-to-one and five-to-five comparisons.
At 40\% or 50\% pruning, it remains competitive, and on par or even better than the bigger ResNet50.
Indeed, the EERs of MobileNetv2 are consistently smaller than ResNet50 (especially in one-to-one comparisons), suggesting that MobileNetv2 is a very robust architecture with capabilities comparable to much larger networks.

We analyze the pruning effect on the networks themselves by plotting learnables and embedding dimensionality (Figure~\ref{fig:results_network_param_size_embedding}).
In all networks, learnables 
decrease slowly until 20-30\% of the filters are removed, then the decrease is sharper, so the filters that are removed first do not affect a high amount of channels.
There is a small exception with SqueezeNet, which has a pronounced drop between 10-15\%, but the decrease stabilizes afterwards.
Embedding size remains constant until certain pruning levels, which, again, is different per network.
ResNet50 starts reducing its output dimensionality with less than 10\% pruning, SqueezeNet at $\sim$20\%, and MobileNetv2 at $\sim$40\%.
A reduction in the embedding indicates that pruning affects the last layer.
Such layer is supposed to be the more specific for the task, so it could be expected that it is not affected in the early pruning stages \citep{caldeira2024arxiv_model_compression_biometrics}.
However, this does not hold with ResNet50.
Nevertheless, ResNet50 is substantially bigger (Table~\ref{tab:networks_used}), its embedding duplicates others (2048 vs. 1280 or 1000), but its performance is not substantially better (Figure~\ref{fig:results_EER_all}).
%
%
Research has shown that feature spaces of 512 or 1024 contain sufficient degrees of freedom to encode a face, allowing further reduction with minor accuracy drops \citep{OTOOLE18TCS_face_representations_DCNN}.
Thus, it is reasonable that the large ResNet50 embedding is affected from the beginning.

\begin{figure*}[htb]
    \centering
    \includegraphics[width=0.95\textwidth]{SQ_MB2_R50_filters_per_layer.png}
    \caption{Effect of pruning in the amount of filters per layer. For each layer, the y-axis indicates the amount of filters between 0\% and 20\% pruning. The circles indicate the layers with the fastest reduction in the amount of filters. The layer names correspond to the networks' release of Matlab r2023b. Better if zoomed.}
    \label{fig:results_filters_per_layer}
\end{figure*}


The latter observations suggest that ResNet50 is over-dimensioned in several aspects. 
At 71\% pruning, it has ten times more parameters than MobileNetv2 (6.97M vs. 0.67M), but not significantly better accuracy: 4.83\% vs. 5.29\% in one-to-one comparisons (8.7\% decrease), or 0.15\% vs. 0.31\% in five-to-five comparisons (51.6\% decrease).
Indeed, at 99\% pruning, ResNet50 learnables are comparable to MobileNetv2 without any pruning.
%
%
To better analyze this effect, we plot in Figure~\ref{fig:results_performance_learnables} the accuracy against the number of learnables.
In one-to-one comparisons, when ResNet50 and MobileNetv2 have similar parameters ($\sim$1.07M), MobileNetv2 has a 2\% smaller EER.
This occurs when MobileNetv2 is pruned at 26\% and ResNet50 at 99\%.
Despite similar depth (conv layers) and parameters, MobiletNetv2 seems \textit{fundamentally} better.
Figure~\ref{fig:_filters_per_layer_26pc_99pc_pruning} shows the filters per layer of MobileNetv2 at 26\% pruning and ResNet50 at 99\%.
ResNet50 layers mostly have less than 150 filters, even in the last stage of bottleneck layers originally designed to increase dimensionality by 4$\times$ (those containing `branch2c' in the name).
%
%
Also, the last part of the network has 250 filters, which contrasts with the 2048 of the unpruned network.
On the other hand, MobileNetv2 maintains a more prominent difference inside bottlenecks, Also, filter count increases substantially towards the end of the network, when the layers are supposed to be more specific for the task, which may explain its superior performance w.r.t. ResNet50.
%
%
%

Interestingly, in the 1-1.1M parameters range, ResNet50 (99\% pruned) and SqueezeNet ($<$10\% pruned) perform similarly, despite the simpler structure of SqueezeNet.
However, the smaller embedding of ResNet50 (245 elements, Figure~\ref{fig:results_network_param_size_embedding}) vs SqueezeNet (1000) suggests that the most intricate structure of ResNet can provide similar performance with a smaller embedding.
On the other hand, differences between ResNet50 and MobileNetv2 disappear with five images per template (bottom plot of Figure~\ref{fig:results_performance_learnables}), compensating for the observed limitations of ResNet50.
It can also be seen that the gap of SqueezeNet remains in five-to-five comparisons, showing the limitations of such a simpler architecture.

We further analyze the effect of pruning 
by plotting (Figure~\ref{fig:results_filters_per_layer}) the filters per layer for different pruning degrees (up to 20\%).
Filters removed first (marked with black circles) belong to layers with the most channels, which are dedicated to dimensionality increase. 
Similarly to the embedding size, these high-dimensional spaces seem over-dimensioned for the task.
%
%
%
This is obviously more prominent in the network with the biggest dimensional spaces (ResNet50).
At 20\% pruning, ResNet50 layers of original dimensionality 2048, 1024, and 512 are reduced to less than 1000, 540, and 255 channels, respectively (less than half).
This aligns with previous findings (Figure~\ref{fig:_filters_per_layer_26pc_99pc_pruning}) showing reduced dimensionality differences inside ResNet50 bottlenecks throughout pruning.
%
%
Thus, it makes sense that high-dimensional channels are affected the most.
%

%
%
%

\begin{figure}[htb]
    \centering
    \includegraphics[width=0.4\textwidth]{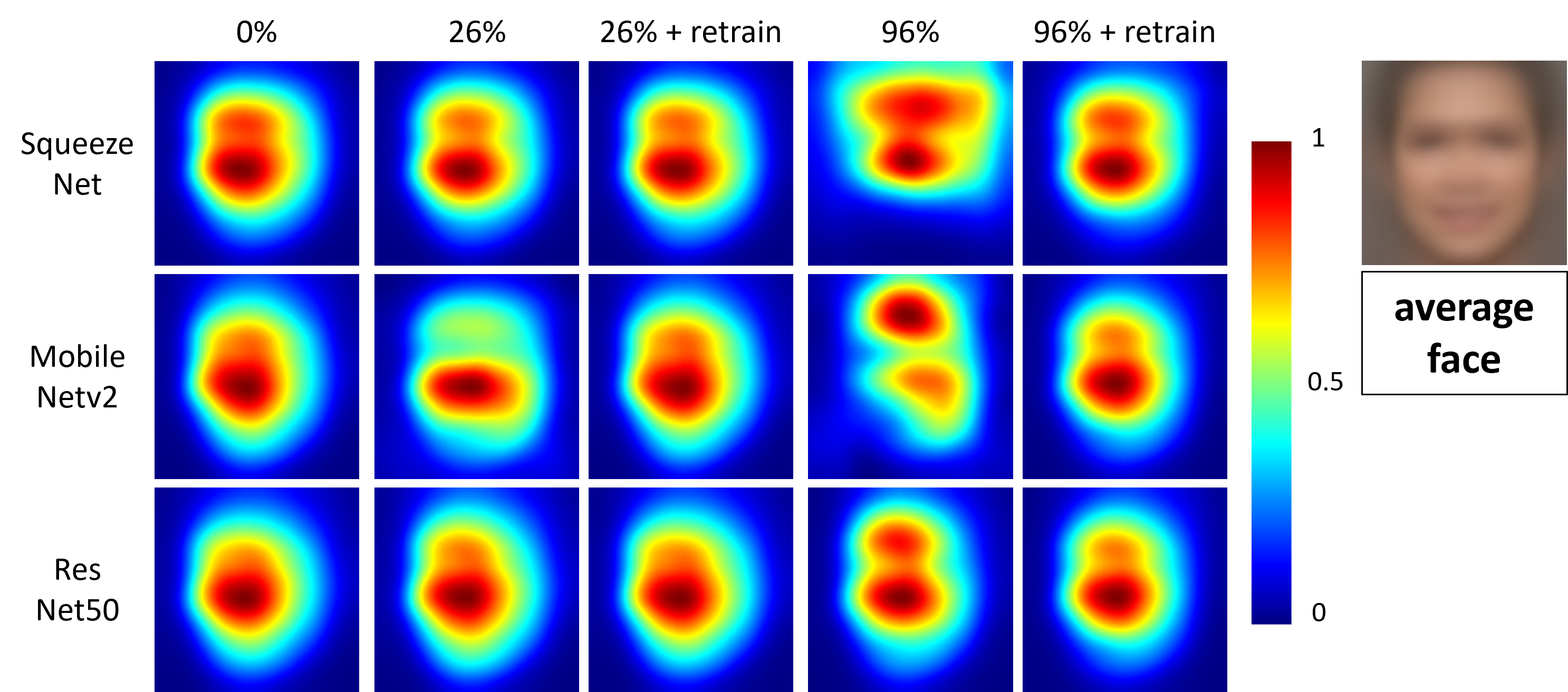}
    \caption{Average heatmaps of the original (unpruned) and different pruned versions of the networks. The average face of VGGFace2-Pose is also shown for reference.}
    \label{fig:results_LIME_heatmaps_db_average}
\end{figure}

\begin{figure*}[htb]
    \centering
    \includegraphics[width=0.95\textwidth]{PSNR_SQ_MB2_R50.png}
    \caption{Histogram of PSNR between the heatmaps of the original (unpruned) and different pruned versions of the networks. The top left of each subfigure shows the one-to-one EER of the unpruned network, whereas the numbers above each curve are the EER of the pruned counterparts.}
    \label{fig:results_LIME_PSNR}
\end{figure*}

\begin{figure}[htb]
    \centering
    \includegraphics[width=0.4\textwidth]{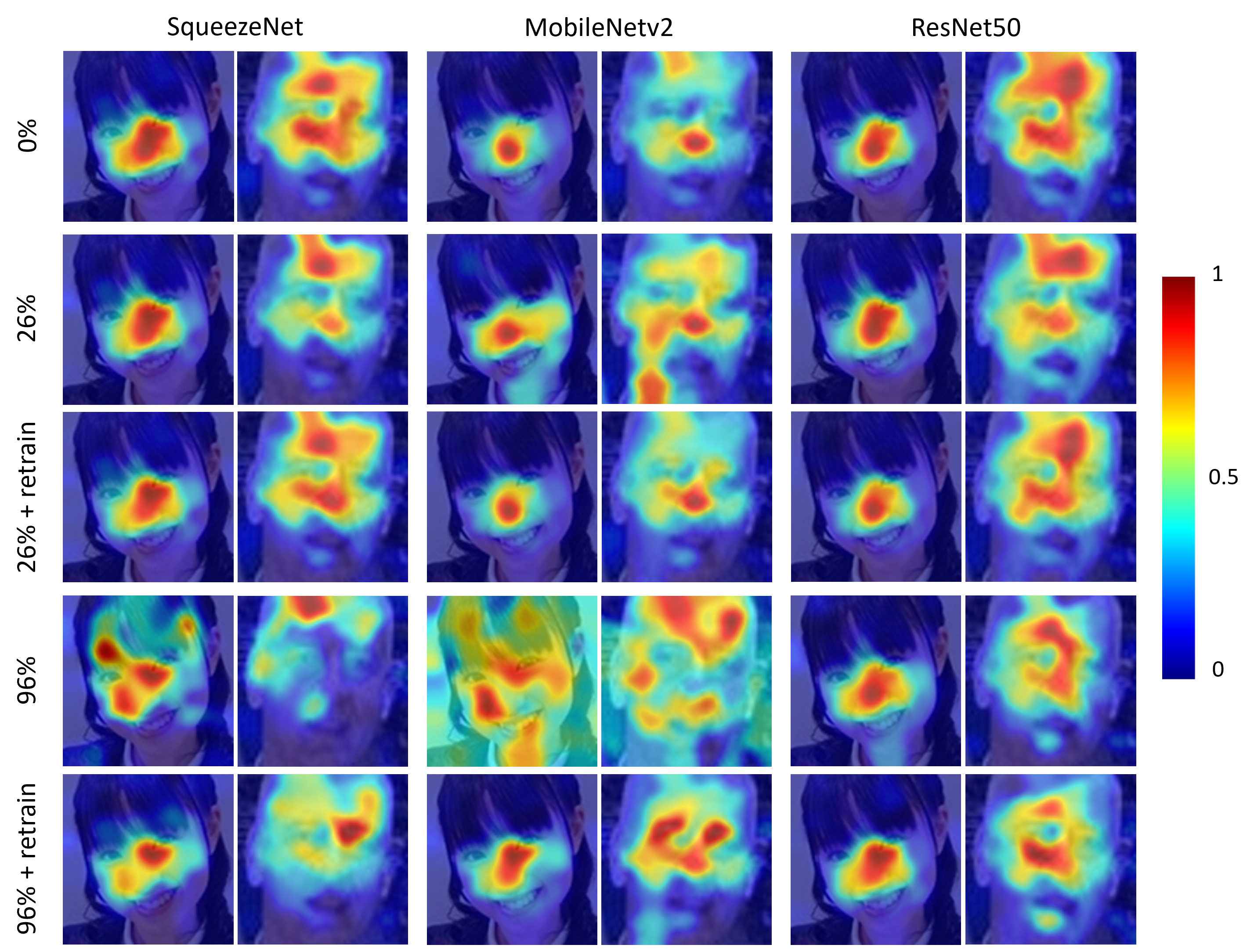}
    \caption{Individual heatmaps of two images of VGGFace2-Pose with the original (unpruned) and different pruned versions of the networks.}
    \label{fig:results_LIME_heatmaps_example_images}
\end{figure}

Finally, we study the impact of pruning on the heatmaps that quantify the most relevant pixels for the networks. We do so via the Local Interpretable Model-Agnostic Explanations (LIME) as employed in \citep{Alonso23wifs_lime_biometrics}.
Figure~\ref{fig:results_LIME_heatmaps_db_average} depicts the average heatmaps of each network on the VGGFace2-Pose set. We show selected pruning levels (26 and 96\%), both with and without retraining. 
As can be seen, the average heatmaps without retraining (columns 2 and 4) depart progressively from the original map of the unpruned networks (column 1). The effect is more remarkable with MobileNetv2, which was observed to be very sensitive to pruning. After retraining (columns 3, 5), the heatmaps recover to a certain extent the shape seen in the first row, which focuses its attention mostly on the nose and eyes area.
We also study the effect at the individual image level by showing in Figure~\ref{fig:results_LIME_PSNR} the histograms of the PSNR between the heatmaps of the original (unpruned) network and the different pruned counterparts. 
As can be expected, retrained networks (solid curves) show higher PSNRs than non-retrained ones (dashed curves). Similarly, pruning at 26\% (red curves) shows higher PSNRs than pruning at 96\% (green curves).
It must be taken into account that at 26 and 96\% pruning, the number of parameters and performance of each network was different (Figure~\ref{fig:results_performance_learnables}). Thus, the ranges of PSNR for each network in the x-axis are different, being closely tied to the EER (shown above each curve). Thus, the more the heatmaps depart from the original (unpruned) version, the higher the EER given by such network.
Lastly, We give some examples of individual heatmaps (Figure~\ref{fig:results_LIME_heatmaps_example_images}) corresponding to the first two users of VGGFace2-Pose.
Similar considerations than with the average heatmaps above can be made. 
At 0\% pruning, the networks focus the attention mostly on the nose area and, to a lesser extent, on the eye area. In one example, the networks also focus on the forehead region.
The heatmaps at 26 and 96\% pruning without retraining (rows 2, 4) depart progressively from the original map of the unpruned networks (row 1). In those cases, the networks sometimes focus their attention on other regions of the image, especially at 96\% pruning. After retraining (rows 3, 5), the networks mostly focus their attention again on the same regions as the original map. 
There are some exceptions, in particular, SqueezeNet and MobileNetv2 at 96\% pruning. However, at this working point, these networks show a degraded EER compared to the unpruned network (Figure~\ref{fig:results_LIME_PSNR}), so we can expect the heatmaps to differ as well.
On the other hand, the EER of ResNet50 does not degrade as much comparatively. Therefore, there is more similarity between heatmaps at different pruning levels.

\section{Conclusions}
\label{sec:conclusions}

Popular face recognition technologies use large Convolutional Neural Networks (CNNs) \citep{[Sundararajan18-DLbiometrics]} with hundreds of megabytes and millions of parameters, which can be impractical for low-capacity hardware.
We are thus interested in methods to achieve lighter CNN architectures without sacrificing performance.
Here, we explore the application of a network pruning method based on Taylor scores, which only requires the backpropagation gradient to measure the impact of filters on classification errors. This allows iterative removal of the least contributing filters.
The method is evaluated on three architectures with varying parameters and depth.
Each network allows a different degree of pruning without substantial performance loss, based on its initial size. For example, the heaviest architecture allows up to 99\% pruning, while the second heaviest allows below 30\%. However, at these points, both have similar learnables ($\sim$1M) and performance.
%
%
Thus, larger networks are more over-dimensioned (at least for the architectures evaluated here).
%
%
%
%
We also observed that filters with the most channels are removed first, regardless of the architecture. This suggests that the high dimensional spaces of the CNNs employed seem to be highly over-dimensioned for the task.
Furthermore, we found that using multiple input images to model the identity of users strengthens the network, allowing higher pruning sparsities.
In the last stage of our experimental setup, we quantify (via heatmaps) the most relevant pixels for the networks. The heatmaps are observed to progressively change with pruning, focusing on regions different from the original unpruned network. After retraining, they somehow resemble the original heatmaps again to an extent that is dependent on the EER gap between the pruned and unpruned networks.

In future work, we are looking at other alternatives for network compression to further optimize the balance between network size and performance, such as pruning combined with Knowledge Distillation (KD). For example, the pruned networks could be trained to mimic the embedding of a more powerful one, e.g. \citep{Deng19CVPR_ArcFace}. However, this would require that both the teacher and student networks have the same embedding size. We are working on this from two perspectives: $i$) by blocking pruning in the last convolutional layer, but also $ii$) by first retraining the original CNNs to have the same output size. The latter approach takes considerably longer, but it will allow the student and teacher to use different architectures.
It has also been suggested that feature spaces of CNN trained in the same task can be linearly mapped \citep{McNeely21arxiv_cnn_embeddings_mapping}, which would solve the issue of embeddings of different sizes.
The combination of several compression techniques, including quantization, KD and pruning, has also shown to be beneficial to achieve model reductions \citep{Rattani23ACCESS_OcularCNNPruningBenchmark}, and it will also be the source of future research.
Finally, we are working on extending this study to other face databases of wide use in the literature \citep{Kolf23ijcb_efar}.

\section*{Acknowledgments}
This work was partly done while F. A.-F. was a visiting researcher at the University of the Balearic Islands.
F. A.-F., K. H.-D., and J. B. thank the Swedish Research Council (VR) and the Swedish Innovation Agency (VINNOVA) for funding their research.
%
This work is part of the Project PID2022-136779OB-C32 (PLEISAR) funded by MICIU/ AEI /10.13039/501100011033/ and FEDER, EU


\bibliographystyle{plainnat}



\end{document}